\documentclass[10pt,conference]{IEEEtran}

\usepackage{cite}

\ifCLASSINFOpdf
   \usepackage[pdftex]{graphicx}
   \graphicspath{{figs/}}
   \DeclareGraphicsExtensions{.pdf,.jpeg,.png}
\else
   \usepackage[dvips]{graphicx}
   \graphicspath{{../figs/}}
   \DeclareGraphicsExtensions{.eps}
\fi
\usepackage[cmex10]{amsmath}
\usepackage{amsthm}

\usepackage{dblfloatfix}

\usepackage{algorithmic}

\usepackage{array}

\ifCLASSOPTIONcompsoc
  \usepackage[caption=false,font=normalsize,labelfont=sf,textfont=sf]{subfig}
\else
  \usepackage[caption=false,font=footnotesize]{subfig}
\fi
\usepackage{url}
\usepackage{todonotes}
\usepackage{amsmath}
\usepackage{amssymb}
\usepackage{physics}

\usepackage{booktabs}
\usepackage{hyperref}

\newcommand{\etal}{\textit{et al}.}

\newcommand{\ie}{\textit{i}.\textit{e}.}
\newcommand{\eg}{\textit{e}.\textit{g}.}

\newif\iffinal
\finalfalse
\finaltrue

\iffinal
\else
\usepackage[switch]{lineno}
\fi

\begin{document}
%

\title{Learning to Detect Good Keypoints to Match \\ Non-Rigid  Objects in RGB Images}


\iffinal

\author{
\IEEEauthorblockN{Welerson Melo$^1$,  Guilherme Potje$^1$,  Felipe Cadar$^1$, Renato Martins$^2$, and Erickson R. Nascimento$^1$\\}
\IEEEauthorblockA{$^1$Universidade Federal de Minas Gerais $^2$Université Bourgogne Franche-Comté\\
{\tt \small \{welerson.melo,guipotje,cadar\}@dcc.ufmg.br, renato.martins@u-bourgogne.fr, erickson@dcc.ufmg.br} }
}


%

\else
  \author{SIBGRAPI paper ID:  44 \\ }
  \linenumbers
\fi

\maketitle

\let\thefootnote\relax\footnotetext{978-1-6654-5385-1/22/\$31.00
	\textcopyright2022 IEEE}

\begin{abstract}
We present a novel learned keypoint detection method designed to maximize the number of correct matches for the task of non-rigid image correspondence. Our training framework uses true correspondences, obtained by matching annotated image pairs with a predefined descriptor extractor, as a ground-truth to train a convolutional neural network (CNN). We optimize the model architecture by applying known geometric transformations to images as the supervisory signal. Experiments show that our method outperforms the state-of-the-art keypoint detector on real images of non-rigid objects by $20$ p.p. on Mean Matching Accuracy and also improves the matching performance of several descriptors when coupled with our detection method. We also employ the proposed method in one challenging real-world application: object retrieval, where our detector exhibits performance on par with the best available keypoint detectors. The source code and trained model are publicly available at \href{https://github.com/verlab/LearningToDetect\_SIBGRAPI\_2022}{https://github.com/verlab/LearningToDetect\_SIBGRAPI\_2022}


\end{abstract}


\IEEEpeerreviewmaketitle


\section{Introduction}
\label{sec:intro}

In the last decades, several methods have been proposed to efficiently detect keypoints. Seminal works such as Harris Corner~\cite{harris1988combined} and SIFT~\cite{lowe2004sift} allowed significant advancements in many applications tasks such as Content-Based Image Retrieval, Structure-from-Motion (SfM)~\cite{heinly2015reconstructing, Potje2017} and registration~\cite{sarlin2020superglue}. 
Keypoint detectors aim to find discriminative visual patterns that are repeatable on different images of the same scene. An effective detector should be invariant to different illumination conditions and equivariant to viewpoint and scale changes. Furthermore, beyond rigid transformations, objects may have different shapes over time due to deformations; therefore, robustness to non-rigid deformations is also a key factor to consider while locating points for visual correspondence. 



Recently, feature detection methods have been moving towards learned-based systems~\cite{laguna2022keynet, revaud2019r2d2, dusmanu2019d2net, luo2020aslfeat, tyszkiewicz2020disk, suwanwimolkul2021learning, vasconcelos2017kvd} delivering results that significantly outperform the handcrafted counterpart~\cite{lowe2004sift, bay2006surf}. However, learning to detect keypoints is not a well-defined task. 
A learned keypoint detector generally learns to find patterns that are repeatable across images; however, this strategy does not guarantee that the detected points are good to be matched. For example, points on edges and repetitive patterns that usually occur in man-made structures are challenging to be matched due to high texture ambiguity. Methods that first detect and then describe keypoints may harm the matching performance in regions where accurate matching is not possible~\cite{revaud2019r2d2}. Moreover, performing matching simultaneously with the detection and description has often high computational complexity~\cite{tyszkiewicz2020disk}. 
Given two images $A$ and $B$ with feature sets $F_A$ and $F_B$, matching them has a time complexity of $O(|F_A|\cdot|F_B|)$. 
As each image pixel may potentially become a feature, the problem quickly becomes intractable, needing carefully designed training schemes and massive computational resources. Regarding the non-rigid transformation, very few works have been proposed to address the non-rigid deformation invariance task. Recent explorations~\cite{you2020ukpgan,nascimento2019geobit,cviu2022-potje} propose to tackle non-rigid deformation but relying on depth information. Despite the advances achieved, RGB cameras are still by far the most common type of imaging sensor.

\begin{figure}[t!]
\vspace{-5pt}
  \centering
    \includegraphics[width=1\linewidth]{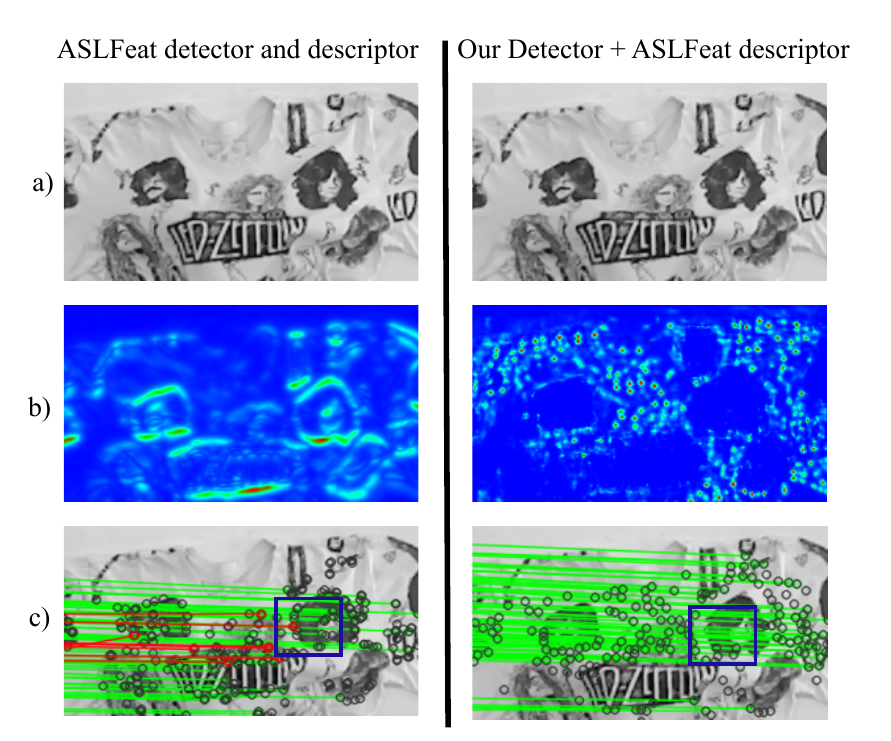}
    \hfill
    \vspace{-20pt}
    \caption{\textbf{Results on detecting good keypoints.} a) Input image with non-rigid deformations; b) Score maps of ASLFet and ours; c) Correct matches (green) and incorrect ones (red), as well as circles representing the detected keypoints. One can notice that not all peaks in the score map are keypoints because we chose the top $1{,}024$ according to the score value. Notably, our method provides more reliable points to be matched.
    }
    \label{fig:teaser}
    \vspace{-10pt}
\end{figure}

In this paper, we present a learned detection strategy designed to tackle non-rigid deformations on still images (Figure~\ref{fig:teaser} shows some qualitative results). We address the keypoint detection problem efficiently in a well-defined manner exploiting the assumption that good features to be detected are also salient points that are likely to yield correct matches. We propose a novel learned detection methodology that predicts ground-truth matching maps based on an existing detector-descriptor configuration. The network is trained to detect good features according to the map derived from true descriptor matches and it can be easily coupled with any combination of existing detector-descriptor pairs. 

We evaluate our detector on three different benchmark datasets (\ie, Kinect1, Kinect2, and DeSurT) of real deformable objects, as well as with application scenarios on content-based object retrieval, validating that our method can reach state-of-the-art performance not only in matching evaluation scores but also in a practical related computer vision task.
Figure~\ref{fig:teaser} illustrates the matching quality of detected keypoints of our detector and the recent ASLFeat detector~\cite{luo2020aslfeat}. 

The contribution of our work is two-fold: (i) a novel keypoint detection training framework aimed to improve the matching performance of existing descriptors, and (ii) the first learned keypoint detector optimized to cope with non-rigid deformations that works only using standard RGB images.

\section{Related work}
\label{sec:related}


\subsection{Handcrafted and learning-based feature detection}
\label{sec:related:sub:handcrafted}
Most of the existing keypoint detectors are based on handcrafted approaches to select feature patterns such as blob, corners or edges~\cite{harris1988combined, lowe2004sift, bay2006surf, rublee2011orb}. 
However, in the past few years, the use of deep learning for feature extraction and image matching pipeline became the new gold standard approach. On the flip side, these learning-based approaches are mostly applied to feature description tasks~\cite{luo2019contextdesc, potje2021extracting} and jointly detection-description~\cite{detone2018superpoint, revaud2019r2d2, dusmanu2019d2net, sarlin2020superglue,luo2020aslfeat, song2020sekd, tyszkiewicz2020disk}.
Few works have focused on learning to detect keypoints by increasing detection probability for repeatable areas between image pairs~\cite{savinov2017quad, zhang2018learning}. In these works, the resulting keypoint detector has high repeatability, although matching performance is hindered since the detected keypoints are often ambiguous~\cite{laguna2022keynet}. Barroso-Laguna~\etal~\cite{laguna2022keynet} use fixed handcrafted filters alongside the learned ones, aiming to improve representation and generalization of low-level features. However, these methods can provide poor detections if the handcrafted filters have some bias that could not help to detect good keypoints, \eg, clustering keypoints along edges and corners. To solve the these problems, the LLF detector~\cite{suwanwimolkul2021learning} trains a detector to retrieve low-level features and keypoint locations by adding a layer that learns to select a keypoint with matching reliability. A drawback of this method is that it does not consider real matching scenarios, which in practice does not improve the matching accuracy of the detected keypoints.

Recently, several works follow the describe-then-detect scheme, where the keypoints are detected from a dense descriptor map. Revaud~\etal~\cite{revaud2019r2d2} claim that detection and description are inseparably tangled, thus keypoints should be detected based on repeatability and reliability. 
Suwanwimolkul~\etal~\cite{suwanwimolkul2021learning} observed that in methods such as D2-Net~\cite{dusmanu2019d2net} and ASLFeat~\cite{luo2020aslfeat}, there is no guarantee that the selected keypoints are associated with matching the learned descriptors, since the keypoint selection is rather handcrafted. Therefore, the matched keypoints do not always have high accuracy. Figure~\ref{fig:teaser} shows these problems of handcrafted keypoint selection in the highlighted squared region. Conversely, in our proposed approach, the peaks are generated directly from the network output (score map in Figure~\ref{fig:teaser}-b), and the detector is trained to improve matching accuracy with real image pairs. 

Fewer works consider the descriptor matching in the training pipeline such as our approach. GLAM~\cite{truong2019glampoints} detects keypoints based on matching quality, however for retinal images, a very specific domain. SEKD~\cite{song2020sekd} proposes a non-domain specific detector and descriptor by first detecting keypoints based on repeatability, and then filtering the reliable keypoints based on the matching. This approach can yield subpar results when a large set of good keypoints for matching is not found in the repeatability optimization stage. DISK~\cite{tyszkiewicz2020disk} considers detection and description in a probabilistic relaxation and applies a reinforcement learning strategy to optimize detection and description jointly. As drawback, the method requires careful hyperparameter tuning to converge. Tonioni~\etal~\cite{tonioni2018learning} trained a decision tree to learn to select 3D keypoints based on good matches. The authors argue that good features to be detected are those likely to yield correct matches. We apply a similar strategy on 2D keypoints. Similarly to GLAM~\cite{truong2019glampoints}, we use the results of matching descriptor with a weight strategy for the Matching Heatmap, applied to a general domain.

\subsection{Non-rigid deformation feature detection and matching}
\label{sec:related:sub:nonrigid}
To circumvent the problem of non-rigid transform and keypoint description, descriptors such as the DEAL method~\cite{potje2021extracting} proposes a deformation-aware local feature description strategy that learns to describe non-rigid patches without depth information. The DaLi descriptor~\cite{moreno2011deformation} encodes features robust to non-rigid deformations and illumination changes. In the same context, GeoBit descriptor~\cite{nascimento2019geobit} uses geodesics from objects surface to compute isometric-invariant features working with RGB-D images. Unlike the proposed approach, these methods are solely concerned with the description and overlook the detection stage. Recently, UKPGAN~\cite{you2020ukpgan} proposes a 3D keypoint detector where keypoints are detected for the task of 3D reconstruction. However, their method is suitable only in context of 3D keypoints, conversely to our proposed detector for 2D images. To the best of the authors' knowledge, no work deals with 2D feature detection on images with non-rigid deformations.  
In this work, we propose a methodology that can be used to obtain keypoints robust to non-rigid deformations relying only on visual information. 
\section{Methodology}
\label{sec:methodology}

\begin{figure}[!t]
\vspace{-5pt}
  \centering
    \includegraphics[width=1\linewidth]{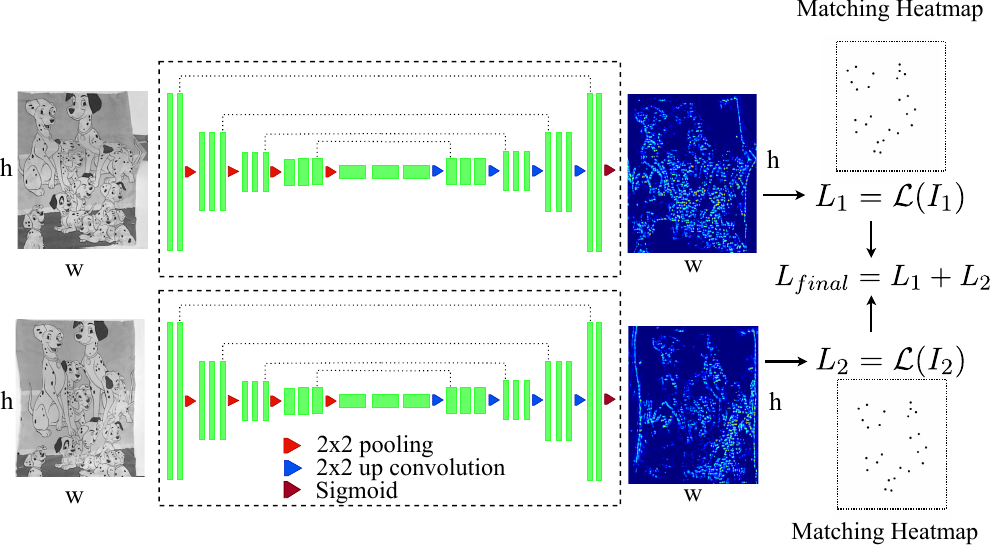}
    \hfill
    \caption{\textbf{Overview of the network architecture used as backbone for keypoint detection.} The Siamese network is optimized to detect reliable keypoints to be matched for a given descriptor. Each green block represents a $3\times3$ 2D convolution layer, followed by batch normalization and ReLU activation function. The two branches share the weights. 
    }
    \label{fig:architecture}
    \vspace{-5pt}
\end{figure}

Our detector is designed to learn to extract keypoints from images being affected by non-rigid deformations. The training setup enforces that keypoints are detected in repeatable locations having confident matching probability for a given descriptor. The learned detector is designed with a 4-level deep U-net~\cite{ronneberger2015unet} with a final sigmoid activation function. It is also composed of $3\times3$ convolution blocks with batch normalization and ReLU activations. U-net is our method of choice due to its past successes in dense regression and semantic segmentation tasks. Figure~\ref{fig:architecture} depicts the network architecture with the Siamese scheme and the final loss $\mathcal{L}$ calculation. In the next sections, we explain the training framework, the loss design, and implementation details. 

\subsection{Keypoint detection learning framework}
\label{subsec:framwork}
Unlike learned keypoints as Key.net~\cite{laguna2022keynet}, and describe-to-detect extraction methods such as R2D2~\cite{revaud2019r2d2}, and ASLFeat~\cite{luo2020aslfeat}, in our learning strategy, the key idea is to leverage an existing detector-descriptor pair to bootstrap the learning process that is focused on high confident matches. Let $A \in \mathbb{R}^{H \times W}$ be an image from our training dataset, defined as the anchor image. We generate images $B$ and $B'$ by applying two different deformations composed of a random homography and a thin-plate spline warp (TPS)~\cite{donato2002approximate} ($g$ and $g'$ respectively) on the anchor image $A$. The TPS warps representing 2D coordinates are often used to model non-rigid deformations, giving us an apparatus to work with this type of transformation. Now, for images $A$, $B$ and $B'$, we detect $k$ salient keypoints according to a base detector. In our experiments, we use $k = 0.02 \times H \times W$, which results in keypoints covering a good portion of all image regions.

\begin{figure}[t!]
\vspace{-5pt}
  \centering\includegraphics[width=0.9\linewidth]{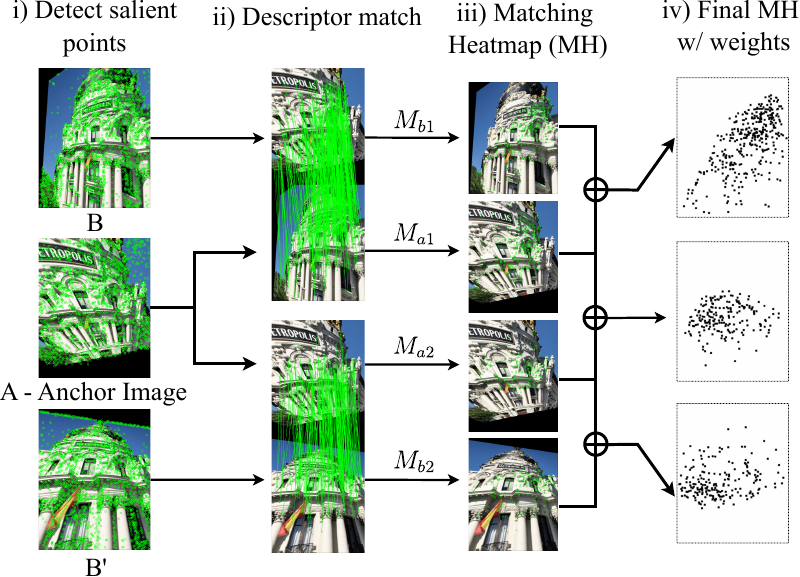}
    \caption{\textbf{Overview of the detector training framework.} Our framework is composed four steps: i) First, we detect the keypoints using a base detector for images A, B and B' (an anchor, and two transformed versions of the anchor with random homography and non-rigid image transformations);
ii) Then, we extract the descriptors on the detected keypoints and then find correspondences with nearest neighbor search;
iii) Using the correct matches, we build a Matching Heatmap (MH) from the location of correct matches for each input image;
iv) MH weighting based on keypoint quality, \ie, true matching repeatability.}
    \label{fig:gt_generation}
    \vspace{-10pt}
\end{figure}

Once we have selected the salient pixels, we extract descriptors for each keypoint location and then match the descriptors of image $A$ with the descriptors of image $B$, and descriptors of image $A$ with descriptors of image $B'$. Please notice that the positions of the correct matches can be found using $g$ and $g'$ in this setup. For each image, the keypoint position $(x,y)$ of a descriptor that passes the mutual nearest and ratio matching tests (and that is also a correct match) is added to the set $C_i$, where $i$ is the index of the keypoint for image $A$, $B$, or $B'$. We train our model to detect keypoints using the location of correct matches of descriptors as ground-truth for the training. We name the generated map using true matches by \textit{Matching Heatmap} (MH). This process is summarised in Figure~\ref{fig:gt_generation}.

Let $M_{a1}, M_{a2}$ be the MH of $A$, and $M_{b1}, M_{b2}$ the MH of $B$ and $B'$ respectively (as can be seen in Figure~\ref{fig:gt_generation}), with values ranging in $[0, 1]$, where the value $0$ means low matching confidence regions and $1$ means high matching confidence regions. The MH has the same resolution of the input image and then we set the MH value as $1$ in the position $(x, y)$ if it is in the set $C_i$. In the last step, we combine the MH from all pairwise matches in a way that map locations have more weight where descriptors were correctly matched on both match attempts, \ie, matches of image $A$ with $B$ and $A$ with $B'$. As a result, we have a final MH for image $A$ as: $M_a = (M_{a1} + M_{a2})/2$. For images $B$ and $B'$, we apply a similar idea, except that now it has three degrees of weights. Considering image $B$, we have descriptors that are correct in both image pairs; descriptors that are correct in the match of $B$ and $A$, and descriptors that are correct in the match of $B'$ and $A$. The latter is also represented on MH of $B$, but with a small weight. The same idea is applied to $B'$. That way, we have the global MHs for $B$: $M_b = (g(M_{a}) + M_{b1})/2$, and for $B'$: $M_{b'} = (g'(M_{a}) + M_{b2})/2$. To make the global MHs easier to be learned by the CNN model, we apply a $3\times 3$ Gaussian kernel in all invidual MHs. 



Finally, the matching map has the information of confident locations to be matched for each image. 
Notice that by choosing only correct matches, we are consequently selecting repeatable keypoints, meaning that our model implicitly learns to be repeatable. To further enforce repeatability of detected points, we also employ a Siamese scheme~\cite{bromley1993signature} to maximize similarities of the score map and the MH of the anchor image and its variations, at same time. These strategies improves the repeatability of the detector under geometric transformations. Our method is agnostic to the choice of the base detector and descriptor. In the experiments section, we will show the capability of the proposed detection approach to improve the matching capability of two recent descriptors.

\subsection{Loss function}
Due to the imbalance between the number of positive and negative pixels in the MH, the full map in the training stage tends to bias the model towards predicting a map with very low scores on average. To solve this problem, we randomly sample a fixed number of  negative examples at each pass of an image in training. Considering that $n$ is the amount of positive examples in an image, we uniformly sample $n$ negatives examples and back-propagate $2n$ examples. We formulate this strategy as a binary pixel-wise mask $F$ having value $1$ on the chosen pixels, and $0$ otherwise. Given an image $I$, its relative MH $M$, and the model output score map $S$, we define $S' = S \times F$. As we aim to maximize the similarity across the MH and the score map image, the cosine similarity ($cossim$) between $S'$ and $M$ is adopted:
\begin{equation}
    \mathcal{L}_{cossim}(I) = 1 - cossim \left(S', M \right).
    \label{eq:loss1}
\end{equation}
When $cossim(S', M)$ is maximized, the MH and the score map tend to be close. Although cosine similarity has good convergence properties, it disregards the magnitude of the values between the score maps and therefore we also consider the L2 loss:
\begin{equation}
    \mathcal{L}_{simple}(I) = \frac{1}{2n}\sum_{i=1}^{H\cdot W}\left(S'_i - M_i\right)^2.
    \label{eq:loss2}
\end{equation}
We further exploit the fact that the regressed map needs to peak at the position of the keypoints. Thus, for even faster convergence, we employ a third loss term in order to force local peakiness of the score map. Considering a set of non overlapping patches $\mathcal{P}=\{p\}$ that contains all $N\times N$ patches within the image $I$ where there is at least one non-zero pixel on the equivalent location of the patch on $M$, the peakiness loss term of the score map is defined as:
\begin{equation}
    \mathcal{L}_{peak}(I) = 1 - \frac{1}{|\mathcal{P}|}\sum_{p \in \mathcal{P}}\left(\max_{(i, j) \in p} S_{i, j} - \underset{(i,j) \in p}{\text{mean}} S_{i,j}\right).
    \label{eq:loss3}
\end{equation}
The final loss $\mathcal{L}$ is given by the weighted sum of the \textit{cossim}, L2 and peak losses:
\begin{equation}
    \mathcal{L}(I) = \lambda_1 \mathcal{L}_{cossim}(I) + \lambda_2 \mathcal{L}_{simple}(I) + \lambda_3 \mathcal{L}_{peak}(I).
    \label{eq:loss_final}
\end{equation}

\subsection{Implementation details}

The weights $\lambda_1$, $\lambda_2$, and $\lambda_3$ were empirically found by performing grid-search on a range of sensible values, and we kept the ones that best enhanced the convergence of the score maps. The weights used in the experiments are $\lambda_1=3.0$, $\lambda_2=1.0$ and $\lambda_3=0.3$. Even though most of the results are from real images, our network is trained using only synthetic warps. We use part of~\cite{potje2021extracting} simulated data to apply the non-rigid deformations and homography as explained on Subsection~\ref{subsec:framwork}. The dataset comprises $400 \times 300$ resolution images. In the train step, a random image from the dataset is chosen as the anchor image $A$ (see Subsection~\ref{subsec:framwork}). In total, $10K$ pairs of images with different and random transformation were used in the training pipeline. We optimize the network via Adam with initial learning rate of $0.006$, scaling it by $0.9$ every $500$ steps for $7$ epochs. We used a batch size of $12$ images containing at least $32$ peaks in its MH. With approximately $150$ positive examples per MH, our model was trained on about $1.5M$ positive examples. 
In the testing, we used non-maximum suppression (NMS) with a window size of $5\times5$ pixels.
We also post-process the keypoints with an edge elimination step such as SIFT~\cite{lowe2004sift} (with a threshold of $10$). The top-k keypoints regarding detection scores are kept, while filtering those whose scores are lower than $0.2$.

\section{Experiments and results}
\label{experiments}

Our approach expects a base detector and a base descriptor to be used for generating the data to our training framework. Given the wide variety of detectors and descriptors available, we chose ASLFeat~\cite{luo2020aslfeat} because in addition to being the state-of-the-art for the detection-description task, its architecture has deformable convolutional kernels. The deformable kernels target to learn dynamic receptive fields to accommodate the ability of modelling geometric variations, which is a nice feature in our context since we are dealing with non-rigid transformations. 
We also selected the DEAL~\cite{potje2021extracting} descriptor that has invariance to non-rigid deformations.
\begin{table*}[t]
\centering
\caption{\textbf{Detector matching performance comparison.} Best in bold and second-best underlined. The higher the value, the better. Results show that our detector provides keypoints on images with non-rigid deformations that enhances the matching results.}
\begin{tabular}{lccccllcccc}
\toprule
\textbf{}                                                                           & \textbf{}   & \textbf{}   & \multicolumn{5}{c}{\textbf{Dataset 770 pairs total - MS / MMA@3 pixels}}                                                                          & \multicolumn{1}{l}{} & \multicolumn{1}{l}{} & \multicolumn{1}{l}{}            \\ \midrule
\multicolumn{1}{c}{\begin{tabular}[c]{@{}c@{}}\textbf{Detector} \\ \textbf{+}\\ \textbf{ASLFeat}\end{tabular}} & Kinect1     & Kinect2     & DeSurT      & \textbf{Mean}                            &  & \multicolumn{1}{c}{\begin{tabular}[c]{@{}c@{}}\textbf{Detector} \\ \textbf{+}\\ \textbf{DEAL}\end{tabular}} & Kinect1     & Kinect2              & DeSurT               & \textbf{Mean}                            \\ \midrule
SIFT                                                                                & 0.35 / \underline{0.77} & 0.37 / \underline{0.85} & 0.26 / \underline{0.63}  & 0.33 / \underline{0.75}                     &  & SIFT                                                                             & 0.33 / \underline{0.68} & 0.38 / \textbf{0.85}          & 0.27 / \textbf{0.63}           & 0.33 / \underline{0.72}                     \\
FAST                                                                                & \underline{0.43} / 0.69 & \textbf{0.53} / \underline{0.85} & \textbf{0.33} / 0.56 & \textbf{0.43} / 0.70                     &  & FAST                                                                             & 0.36 / 0.58 & \textbf{0.51} / \underline{0.81}          & \textbf{0.29} / 0.49          & \underline{0.39} / 0.63                  \\
AKAZE                                                                               & 0.39 / 0.66 & \underline{0.49} / 0.76 & 0.26 / 0.48 & \underline{0.40} / 0.66                     &  & AKAZE                                                                            & \underline{0.38} / 0.65 & \underline{0.47} / 0.74          & 0.23 / 0.42          & 0.36 / 0.60                     \\
Keynet                                                                              & 0.31 / 0.65      & 0.35 / 0.62        & 0.24 / 0.51        & 0.30 / 0.59                            &  & Keynet                                                                           & 0.27 / 0.58 & 0.34 / 0.59           & 0.22 / 0.45           & 0.28 / 0.54                      \\
ASLFeat                                                                             & 0.31 / 0.58 & 0.39 / 0.69 & 0.28 / 0.53 & 0.33 / 0.60                     &  & ASLFeat                                                                          & 0.31 / 0.66     & 0.40 / 0.73              & 0.25 / 0.54              & 0.32 / 0.64                         \\ \midrule
Ours                                                                                & \textbf{0.49} / \textbf{0.86} & 0.48 / \textbf{0.89} & \underline{0.31} / \textbf{0.66} & \multicolumn{1}{l}{\textbf{0.43} / \textbf{0.80}} &  & Ours                                                                             & \textbf{0.45} / \textbf{0.79} & 0.46 / \textbf{0.85}          & \underline{0.28} / \underline{0.59}          & \multicolumn{1}{l}{\textbf{0.40} / \textbf{0.74}} \\ \bottomrule
\end{tabular}
\label{tab:results_match}
\end{table*}

We evaluate our detector in different publicly available datasets containing deformable objects in diverse viewing conditions such as illumination, viewpoint, and deformation. For that, we have selected the two datasets recently proposed by GeoBit~\cite{nascimento2019geobit,cviu2022-potje} and one by DeSurT~\cite{wang2019deformable}. They contain color
images of $11$ deforming real-world objects.

\subsection{Metrics and baselines}
\label{experiments:metrics}

Since the main goal of our feature detection is to maximize the number of correct feature matches, the main performance assessment is done with the Mean Matching Accuracy (MMA) as in Revaud~\etal~\cite{revaud2019r2d2}, which computes the ratio of correct matches and possible matches; and the 
Matching Score (MS)~\cite{revaud2019r2d2}. We also use the keypoint Repeatability Rate (RR), the most used keypoint metric, defined as the ratio of possible matches and the minimum number of keypoints in the shared view with a $3$ pixels error threshold. 


We compare our results against five detectors. We consider two handcrafted detectors: SIFT~\cite{lowe2004sift} and AKAZE~\cite{alcantarilla2012kaze}, which provides stable keypoints and are still considered good baselines according to a recent study~\cite{jin2021image}; FAST~\cite{rosten2006fast}, a basic corner detector; Keynet~\cite{laguna2022keynet}, a cutting edge learning-based detector; and one state-of-the-art jointly learned detection and description method, ASLFeat~\cite{luo2020aslfeat}. 

\subsection{Results on sequences of deformable objects}
\label{experiments:results}

\begin{figure}[t!]
	\centering
	\includegraphics[width=1\linewidth]{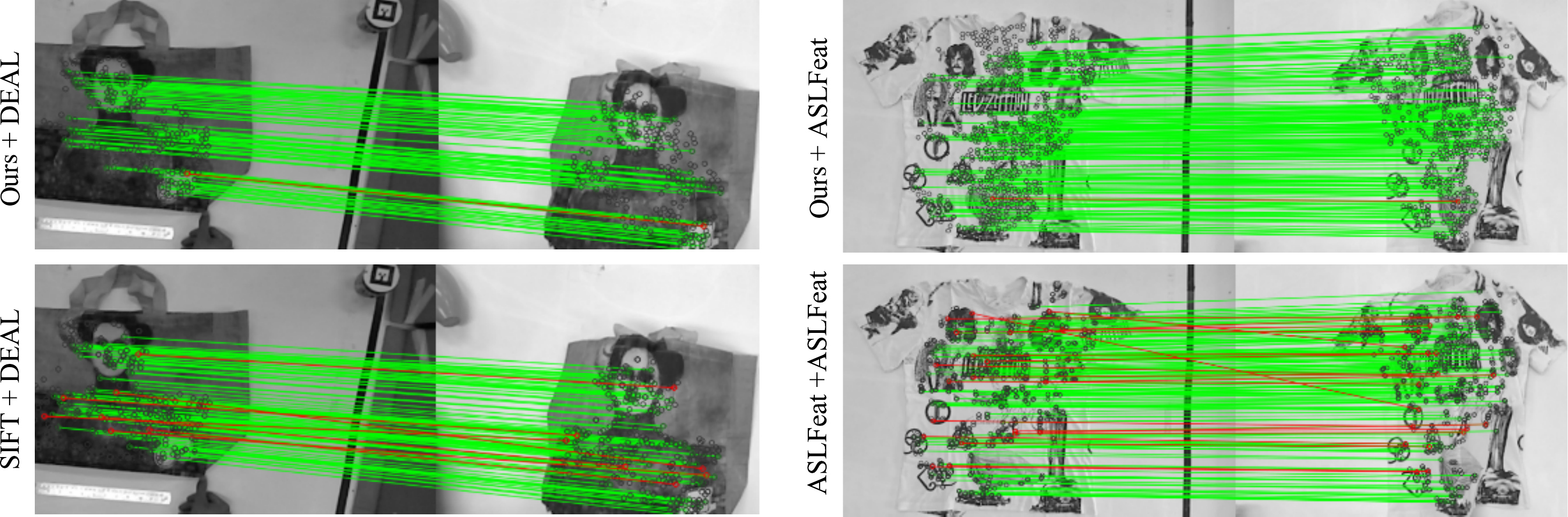}
	\caption{\textbf{Qualitative results on a real non-rigid matching example.} The green lines show correct correspondences, while red lines depict wrong correspondences. Our keypoint detector improves matching accuracy of DEAL and ASLFeat descriptors with respect to SIFT and ASLFeat detectors, while maintaining a similar number of total matches.  }
	\label{fig:qualitative}
\end{figure}

Table~\ref{tab:results_match} shows the results of the experiments using the keypoint detectors for matching image pairs. We detect $1{,}024$ points for each detector on each image. Our main experiment aimed to analyze how the detected keypoints influence the quality of the matching. For this purpose, we chose two descriptors: DEAL~\cite{potje2021extracting} and ASLFeat~\cite{luo2020aslfeat}.
One can see that, on average, our keypoints paired with DEAL descriptors outperforms all detector-DEAL combinations in both MS and MMA metrics. It is also clear that for the ASLFeat descriptor, our detector reaches the best MMA for all datasets. It is worth mentioning that our method increases the avg. MMA  scores from $0.60$ to $0.80$ ($20$ p.p.) when replacing ASLFeat's detector with our detector, and has a significant distance of $5$ p.p. from the second best MMA (SIFT-ASLFeat). For the DEAL descriptor, our detector achieves most of the best and second-best MS and MMA scores, increasing, on average, about $7$ p.p. and $2$ p.p. for MS and MMA, respectively, in comparison with SIFT detector used to train the DEAL descriptor. Figure~\ref{fig:qualitative} shows a matching example of our detector combined with different descriptors. Our method is able to deliver well-distributed matches in the image as well as the SIFT and ASLFeat detectors, but with improved accuracy.

We also analyze the detectors' repeatability with the RR scores. Of all methods, FAST has the best RR with $0.59$ in average. As second best, AKAZE, ASLFeat, and our method reach a RR of $0.50$. The detector with the worst RR metric was SIFT with $0.43$. These scores show that our detector could also reach a competitive RR, while increasing the MMA and MS matching metrics. It is worth to notice that the ASLFeat detector, even if presenting a high RR has the smaller MS and MMA, as can be seen in Table~\ref{tab:results_match}. One can also note that a high RR does not imply good matching scores.

\subsection{Results on the application of object retrieval} 

\begin{figure}[t!]
\vspace{-10pt}
  \centering
    \includegraphics[width=1\linewidth]{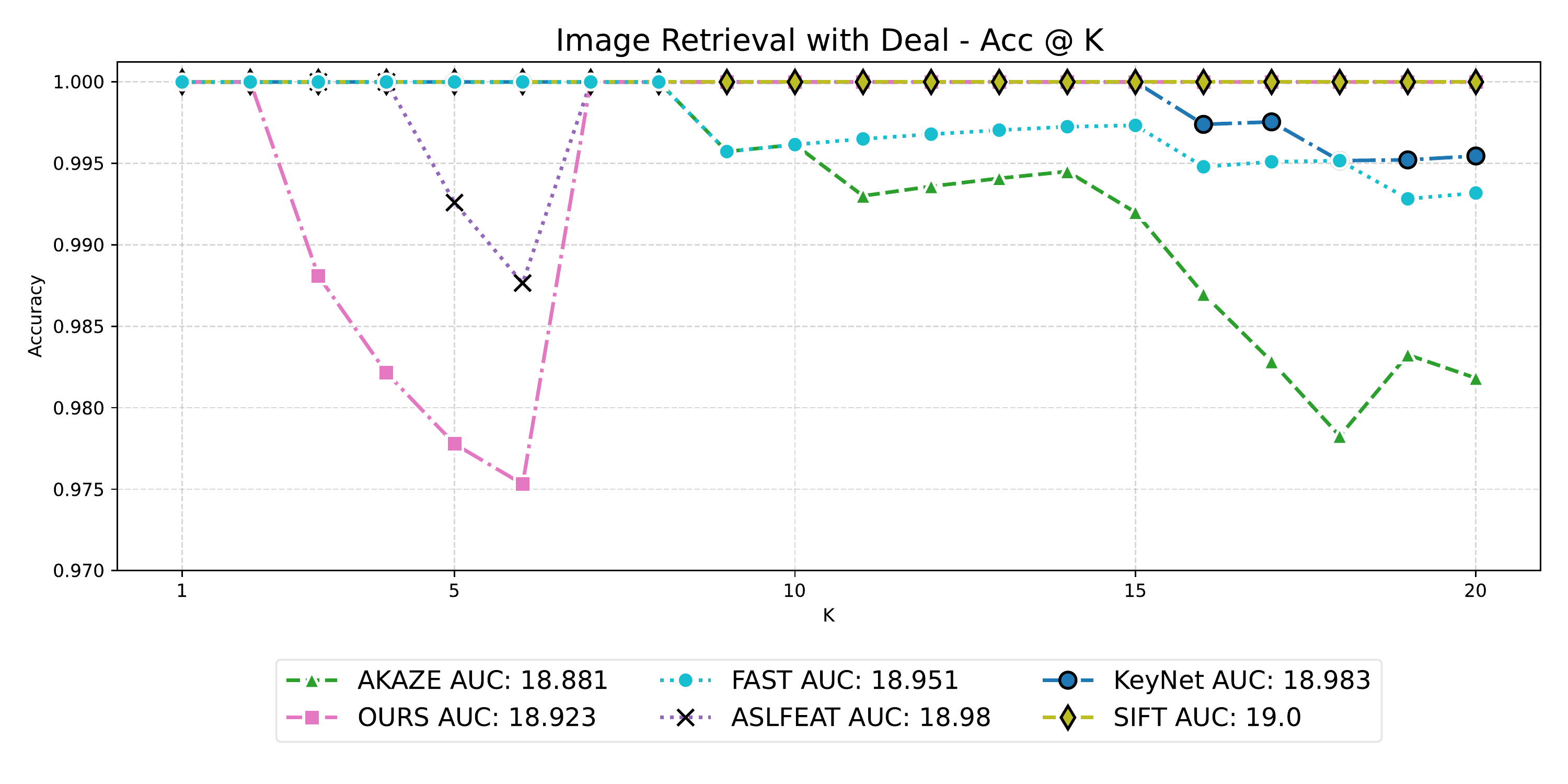}
    \hfill
    \vspace*{-0.7cm}
    \caption{\textbf{Non-rigid object retrieval application.} The graph shows the retrieval accuracy@K for $K=20$ using a nonrigid descriptor and various detectors.}
    \label{fig:retrieval}
\end{figure}

To further demonstrate the effectiveness of our detector in potential applications, we performed experiments in one important related real-world task: content-based object retrieval. The goal is to retrieve the top K images corresponding to a given query. To represent each image, we used a Bag-of-Visual-Words approach. For each keypoint, we first construct a visual dictionary with the DEAL~\cite{potje2021extracting} descriptor, which is used to compute a global descriptor for each image. Given a query image, we calculate the global descriptor and use the K-Nearest Neighbor search to obtain the top K closest objects. We use the same datasets of Table~\ref{tab:results_match}. We use retrieval accuracy (the number of correct objects retrieved in the top K images) to evaluate the performance of the detectors. Since the queries and database of the application are deformable, we choose only to use a descriptor that models isometric deformations. Figure~\ref{fig:retrieval} shows the retrieval accuracy for $K=20$, where our detector performed similarly to the other methods. For $K>6$,  our detector performed similar to SIFT, with is the method used to train the nonrigid descriptor. The results indicate that our detector can perform well even on a non-matching task.

\subsection{Ablation and sensitivity analysis}
\label{experiments:ablation}

As part of ablation studies, we train our model using two configurations: (i) a Siamese network scheme and (ii) using a standard network training scheme, \ie, using a single branch. The experiments show that a Siamese scheme (i) helps the model to learn repeatable keypoints and improve MS. RR increased from $0.46$ to $0.50$, and MS from $0.39$ to $0.43$ by using the Siamese scheme, and MMA decreased from $0.81$ to $0.80$, however inliers significantly increased from $150$ to $170$. 

To evaluate the contribution of the components of our proposed loss (Equation~\ref{eq:loss_final}), and support our implementations decisions, we evaluate three different setups: (i) using cosine similarity term only; (ii) full loss of Equation~\ref{eq:loss_final} with equal weights to cosine similarity and L2 losses, \ie, $\lambda_1=1.0$, $\lambda_2=1.0$, and $\lambda_3=0.3$; and (iii) the \textit{complete loss} of Equation~\ref{eq:loss_final} with optimal weights. 
The results in Table~\ref{table:ablation} show that setup (iii) is the best one. MMA@3 with a value of $1$ p.p. higher for the setup (i) can be explained by the smaller number of inliers. \textit{Complete loss} setup has a higher number of inliers and MS, maintaining a high MMA@3. 
To support the weighting step choice in our training framework, we also report the strategy of giving equal weights according to repeatable matching, where all MHs peaks has a constant value of $1.0$. 
We obtained values of $0.45$ and $0.37$ for the RR and MS on equal weights strategy, which is significantly lower than what we achieve using the proposed weighted Matching Heatmap strategy ($0.50$ and $0.43$ for RR and MS, respectively).

\begin{table}[t!]
\centering
\vspace*{-0.7cm}
\caption{\textbf{Ablation Analysis.} The higher the better.}
\begin{tabular}{lcccc}
\toprule
\textbf{Loss combination} & \textbf{RR} & \textbf{MS} & \textbf{MMA@3} & \textbf{Inliers} \\ \midrule
 $\mathcal{L}_{cossim}$                        & 0.42             & 0.37             & \textbf{0.81}       & 121                   \\
$\mathcal{L}_{cossim} + \mathcal{L}_{simple}$                        & 0.48             & 0.39             & 0.80                & 155                   \\
$\mathcal{L}$ (complete)                       & \textbf{0.50}    & \textbf{0.43}    & 0.80                & \textbf{170}          \\ \bottomrule
\vspace{-10pt}
\end{tabular}
\label{table:ablation}
\end{table}

\section{Conclusion}
\label{conclusion}
We introduce a novel approach to detect keypoints on images affected by non-rigid deformations, emphasizing improved matching scores. To that end, we proposed a training framework for training a CNN with non-rigidly deformed images exploring the hypothesis that a detector can learn the likelihood of correct matching for a given descriptor. The experimental results show  that our method achieved the state-of-art detection and matching performance on non-rigid deformation datasets. Through extensive investigation, we observed that the repeatability of the detector alone is not enough to make a good detector. We also show the efficiency of our detector in a real-world application, demonstrating that learning to detect good keypoints is a promising research direction for performance improvement in real-world tasks. A limitation of our work is that the framework still depends on a base keypoint detector, and may be biased toward specific local characteristics of the base detector. As a future work direction, we plan to investigate how to remove the base detector from the pipeline and learning to detect directly from the descriptor.


\section*{Acknowledgments}
The authors would like to thank CAPES, CNPq, FAPEMIG, Google and Petrobras for funding different parts of this work. We also thank NVIDIA for the donation of a Titan XP GPU.

\bibliographystyle{IEEEtran}
\bibliography{egbib}
%
%


\end{document}